%% file: main.tex
\documentclass[lettersize,journal]{IEEEtran}
\usepackage{amsmath,amsfonts}
\usepackage{algorithmic}
\usepackage{array}
\usepackage[caption=false,font=normalsize,labelfont=sf,textfont=sf]{subfig}
\usepackage{textcomp}
\usepackage{stfloats}
\usepackage{url}
\usepackage{verbatim}
\usepackage{cite}
\usepackage{pifont}
\usepackage{bbm}
\usepackage{amsmath}
\usepackage{amssymb}
\usepackage{booktabs}
\usepackage{multirow}
\usepackage{xcolor}
\usepackage{url}
\usepackage{caption}
\usepackage{subcaption}

\usepackage{colortbl}
\definecolor{mygray}{rgb}{1,0.95,0.97}

\usepackage{xcolor}
\definecolor{LightCyan}{RGB}{0.88,1,1}
\definecolor{sgreen}{RGB}{30, 150, 30} 
\definecolor{cvprblue}{RGB}{54,125,189}

\usepackage[pagebackref,breaklinks,colorlinks,allcolors=blue]{hyperref}

\usepackage[capitalize]{cleveref}
\crefname{section}{Sec.}{Secs.}
\Crefname{section}{Section}{Sections}
\Crefname{table}{Table}{Tables}
\crefname{table}{Tab.}{Tabs.}

\usepackage{bm}
\usepackage{amsthm,amsmath,amssymb}
\usepackage{mathrsfs}
\usepackage{booktabs}
\usepackage{multirow}
\usepackage{multicol}
\usepackage{float}
\usepackage{subfig}
\usepackage[table]{xcolor}
\usepackage{multirow}
\usepackage{enumitem}
\usepackage{bbding}
\usepackage{mathrsfs}
\usepackage{color}
\usepackage{colortbl}
\usepackage{rotating}
\usepackage{array}

\usepackage{xcolor}
\usepackage{adjustbox}

\usepackage{tabularx}
\usepackage{etoolbox}
\usepackage{makecell}
\usepackage{pifont}

\usepackage{wrapfig}
\usepackage{cutwin}
\usepackage{alltt}
\usepackage{graphicx}
\usepackage{multirow}
\usepackage{enumitem}
\usepackage{bbding}

\usepackage{mathrsfs}
\usepackage{booktabs}
\usepackage{multicol}
\usepackage{color}
\usepackage{float}
\usepackage{subfig}
\usepackage{colortbl}
\usepackage{rotating}
\usepackage{array}

\usepackage[ruled]{algorithm2e}

\SetAlgoNlRelativeSize{0}

\begin{document}

\title{MooD: Perception-Enhanced Efficient \\ Affective Image Editing via Continuous \\ Valence–Arousal Modeling}

\author{
Xinyi Yin, Yiduo Wang, Tingqi Hu, Meicong Si, Yunyun Shi, Shi Chen, Hao Wang, Junxiao Xue, Xuecheng Wu

\thanks{Received 13 May 2026; revised 9 August 2026 and 12 September 2026; accepted 17 September 2026. This work was supported in part by the Central Government Guiding Local Science and Technology Development Fund (Grant No. 2026ZY04001), the 2026 AI Breakthrough Initiative Research Project (Grant No. 2026JLGJ0001GX), as well as the National Natural Science Foundation of China (Grant No. 12671458).}
\thanks{Xinyi Yin, Yiduo Wang, and Meicong Si are currently with the School of Cyber Science and Engineering, Zhengzhou University, Zhengzhou 450002, China. (E-mail: {\small yinxinyi@stu.zzu.edu.cn})}
\thanks{Tingqi Hu is currently with the Institute of Information Engineering, Chinese Academy of Sciences, Beijing 100085, China. (E-mail: {\small hutingqi26@mails.ucas.ac.cn})}
\thanks{Yunyun Shi and Shi Chen are currently with the School of Computer Science and Technology, Xi'an Jiaotong University, Xi'an 710049, China. (E-mail: {\small 
yunyunshi@stu.xjtu.edu.cn})}
\thanks{Hao Wang is with the School of Journalism and New Media, Xi'an Jiaotong University, Xi'an 710049, China. (E-mail: {\small haowangx@xjtu.edu.cn})}
\thanks{Junxiao Xue is with the Research Center for Space Computing System, Zhejiang Lab, Hangzhou 311100, China. (E-mail: {\small xuejx@zhejianglab.cn})}
\thanks{Xuecheng Wu is with the Department of Electronic Engineering \& MMLab, The Chinese University of Hong Kong, Sha Tin, Hong Kong SAR, China. (E-mail: {\small wuxc3@ieee.org});}
\thanks{Xinyi Yin and Yiduo Wang deserve equal contributions.}
\thanks{Corresponding author: Xuecheng Wu.}
}

\markboth{IEEE Transactions on Computational Social Systems}
{Shell \MakeLowercase{\textit{et al.}}: A Sample Article Using IEEEtran.cls for IEEE Journals}

\maketitle

\input{Sec/0_abs}

\vspace{-1.8em}

\input{Sec/1_intro}
\input{Sec/2_related}
\input{Sec/3_method}
\input{Sec/4_expers}
\input{Sec/5_conclusions}

\bibliographystyle{IEEEtran}
\bibliography{main}

\end{document}

%% file: Sec/0_abs.tex
\begin{abstract}
Affective Image Editing (AIE) aims to modify visual content to evoke targeted emotions. Although current approaches achieve impressive editing quality, they often overlook inference efficiency, which limits their applicability in computational social scenarios. Moreover, most methods depend on discrete emotion representations, which hinder the continuous modeling of complex human emotions and constrain expressive capabilities in interactive scenarios. To tackle these gaps, we propose \textbf{MooD}, the first VA-driven framework that directly leverages continuous Valence–Arousal (VA) values as editing instructions for fine-grained and efficient AIE in computational social systems. Specifically, we first introduce a VA-Aware Retrieval strategy to bridge vague affective values and detailed visual semantics. Building upon this, MooD integrates visual transfer and perception-enhanced semantic guidance to achieve controllable AIE. Furthermore, considering that existing VA-annotated datasets mainly focus on social scenarios and largely overlook natural scenes, we therefore construct AffectSet, a comprehensive VA-annotated dataset covering diverse scenarios, to support model optimization and evaluation. Extensive qualitative and quantitative experimental results demonstrate that our MooD achieves superior performance in both affective controllability and visual fidelity while maintaining high efficiency. A series of ablation studies further reveals the crucial factors of our introduced design. Our code is available at \href{https://github.com/XinyiYin/MooD}{GitHub}.
\end{abstract}

%% file: Sec/1_intro.tex
\section{Introduction}
\label{sec:introduction}

\IEEEPARstart{O}{n} the currently popular online social platforms, user-generated content (UGC) can convey rich emotional information that directly affects user engagement, social behaviors, as well as interaction quality~\cite{wu2025avf,wu2025towards}, making effective emotion recognition and analysis a critical research focus~\cite{tcss_1, tcss_3, add_4}. Building upon these capabilities, affective attributes can be modulated to enhance recommendation performance, for example by tailoring content to users' emotional states while protecting their preference data~\cite{add_2, add_3}. Beyond recommendation systems, such affective modulation also benefits various social applications, such as public opinion analysis~\cite{social}, human-computer interaction~\cite{human_computer_interaction}, and healthcare~\cite{tcss_2}. This has given rise to the task concept of Affective Image Editing (AIE), which aims to guide viewers toward targeted emotions through visual editing, thereby potentially influencing engagement and personalization in recommendation feeds. However, the large-scale and open-domain nature of UGC poses challenges for real-time, efficient AIE, highlighting the need for fast and practical frameworks deployable in social systems.

\begin{figure}[t!]
\centering
\includegraphics[width=\linewidth]{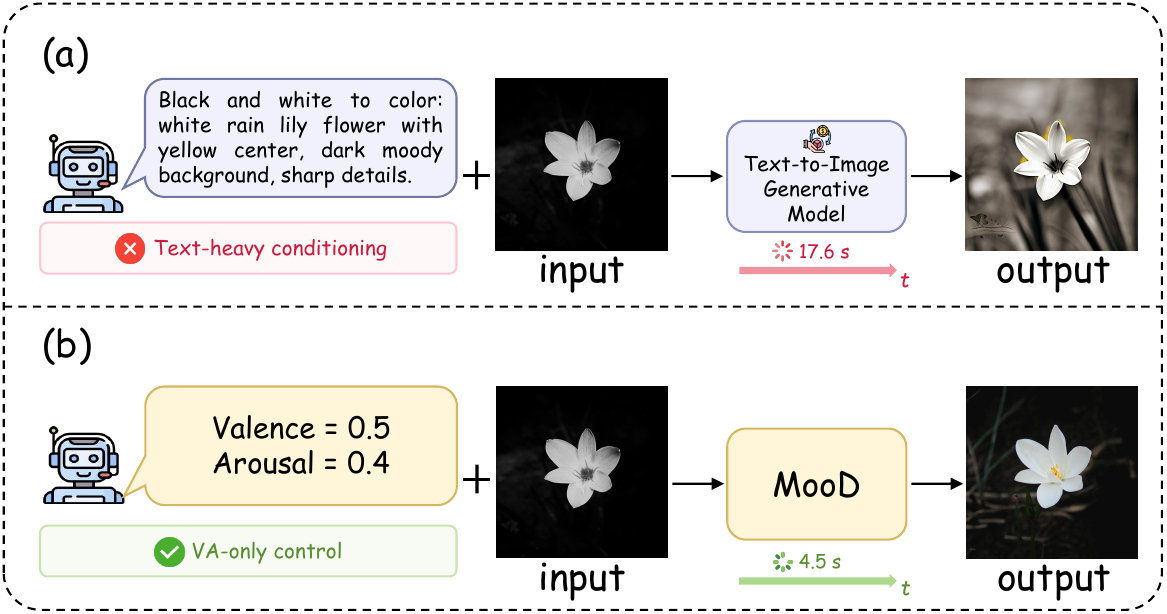}
\vspace{-1em}
\caption{The comparisons between \textbf{(a)} prompt-based AIE methods (\textit{e.g.}, ControlNet \cite{controlnet}) requiring lengthy descriptive inputs and \textbf{(b)} our MooD framework, which takes pure VA values as direct input for continuous AIE.}
\vspace{-1.8em}
\label{fig:motivation}
\end{figure}

Despite the growing interest in AIE, current research still primarily focuses on improving generation quality. Early works mainly relied on low-level visual adjustments such as color  transfer~\cite{color_transfer_3} and style transfer~\cite{style_transfer_1, clva, style_transfer_3} to alter the emotional tone of images. Benefiting from the impressive development of diffusion-based text-to-image models~\cite{sd15,sdxl}, recent methods~\cite{emoedit,emokgedit,make_me_happier} have extended image editing to more advanced semantic-level control. Moreover, with advances in large language models (LLMs) and multimodal large language models (MLLMs)~\cite{xue2026towards,wu2025vic,zhang2025hkd4vlm}, some methods~\cite{emoagent,towards_LLM} have incorporated these models into AIE. Despite their promising results, these methods typically rely on computationally intensive inference pipelines, such as multi-round iterative optimization~\cite{make_me_happier} and MLLM-based agentic reasoning~\cite{emokgedit,emoagent,towards_LLM}. As a result, they lead to increased latency and substantial computational overhead, ultimately limiting the scalability and efficiency of AIE.

\begin{figure}[t!]
\centering
\includegraphics[width=0.8\linewidth]{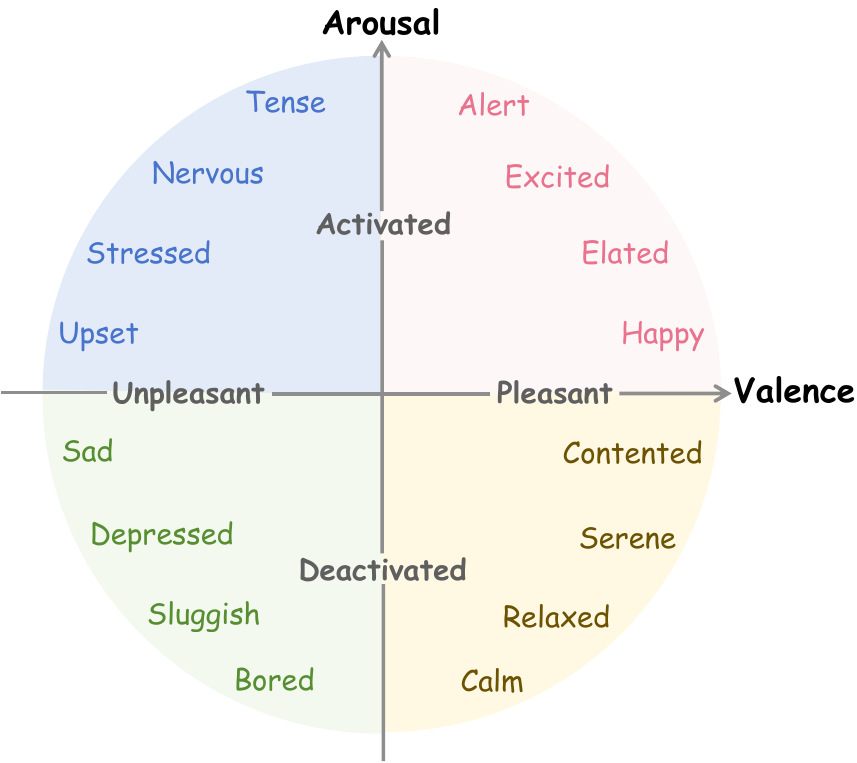}
\vspace{-0.2em}
\caption{The overall illustration of the Valence-Arousal space.}
\vspace{-2em}
\label{fig:VA_Space}
\end{figure}

Beyond inference efficiency, existing AIE methods also suffer from limitations in emotional representation modeling, as most of them depend on categorical emotion state (CES). It is acknowledged that human emotions are inherently continuous and blended~\cite{circumplex_model}. In this situation, such rigid categorical boundaries fail to capture subtle transitions and fine-grained variations in emotional states. Therefore, advancing AIE that can capture fine-grained emotions remains a key challenge. A central consideration in overcoming this challenge is the method of emotion modeling. Using the Dimensional Emotion Space (DES) in psychology (Fig.~\ref{fig:VA_Space})~\cite{circumplex_model}, some methods~\cite{emoticrafter} have adopted the Valence-Arousal (VA) space to represent emotions continuously, which has demonstrated promising performance. Motivated by this observation and the demand for intuitive user interaction, a natural question arises: can VA values be directly utilized as control signals to achieve fine-grained and efficient AIE (Fig.~\ref{fig:motivation})? The answer is affirmative in principle, as VA values provide compact and continuous representations of emotions. However, a key challenge lies in bridging VA values with concrete visual semantics.

Unlike textual instruction inputs, directly employing VA values for AIE is challenging due to the semantic gap between abstract VA values and concrete visual operations. For example, it is unclear what visual changes should correspond to $(V, A) = (0.7, 0.3)$. To bridge this gap, we propose retrieving reference images from annotated affective datasets that match the target VA values, and injecting their affective features into the source image via a lightweight visual adapter, thereby providing explicit semantic guidance. However, this solution introduces new challenges. The adapter exhibits perception defects that can cause it to misinterpret or overly generalize visual features from reference images, resulting in artifacts such as incorrect structures or layouts, as well as hallucinated content and semantic contamination. Large-scale text-to-image models such as Stable Diffusion XL~\cite{sdxl} are trained on massive amounts of image-text pairs and thus encode rich visual prior knowledge. However, their generation process heavily relies on textual conditions to provide semantic constraints. Weak textual emotion cues make the generation insufficiently constrained, leading to excessive reliance on reference images and resulting in uncontrolled transfer.

Based on the above analysis, we formulate three key principles for a VA-controllable and efficient AIE method: \textbf{(1)} Reference images should ensure both emotional accuracy and scene compatibility to maintain editing quality;
\textbf{(2)} The diffusion-based text-to-image model should be guided by perception-enhanced explicit editing priors to enable selective absorption of emotion-relevant visual semantics from reference images; \textbf{(3)} The inference framework should remain lightweight to overcome the efficiency bottleneck of existing methods.

Building upon these principles, we propose \textbf{MooD}, a VA-driven framework for efficient and controllable AIE that achieves fine-grained emotional control in social systems. Specifically, we first introduce a VA-Aware retrieval strategy that enforces emotion–semantic consistency, ensuring both emotional accuracy and visual coherence. Second, we develop a dual-channel editing pipeline from the complementary perspectives of visual transfer and perception-enhanced semantic guidance. The image adapter serves as the visual channel, enabling visual feature transfer, while we further propose an Affective Semantic Projection Network (ASP-Net) as the semantic channel. ASP-Net provides perception-enhanced guidance to achieve fine-grained semantic control by projecting emotion-relevant visual features from the reference image into a CLIP text embedding space aligned with the source image, transforming unconstrained feature injection into a semantically guided process. Through the synergy of this dual-channel design, our MooD effectively suppresses hallucinated structures and semantic contamination while preserving the target emotional expression.

Furthermore, since existing VA-annotated datasets focus mainly on social scenarios and largely overlook natural scenes, we construct a comprehensive visual affective dataset with a total of 63,387 samples termed AffectSet by selecting samples with object attribute labels from EmoSet~\cite{emoset} and annotating them with VA values utilizing Qwen2.5-VL-72B~\cite{Qwen2.5-VL}. Notably, AffectSet achieves Pearson correlation coefficients of 0.914 and 0.837 with few-shot human annotations on the Valence and Arousal dimensions, respectively, demonstrating the reliability of its labels. Extensive qualitative and quantitative experimental results on the inference set proposed by EmoEdit~\cite{emoedit} demonstrate the effectiveness of our MooD. A series of ablation studies is also conducted to indicate the crucial design factors behind model performance. In summary, the main contributions of this paper are three-fold:

\begin{itemize}

 \item We propose \textbf{MooD}, the first efficient VA-driven framework that conditions affective image editing directly on continuous VA values for fine-grained emotional control. We further design a lightweight dual-channel editing pipeline that achieves efficient inference while preserving editing quality.

\item We design a VA-Aware retrieval strategy for emotion-aligned image retrieval, and an Affective Semantic Projection Network (ASP-Net) that maps emotion-relevant visual features into CLIP text embeddings conditioned on source images, offering perception-enhanced semantic guidance for diffusion-based editing.

\item We construct a VA-annotated visual affective dataset termed AffectSet to support model optimization and evaluation. Extensive quantitative and qualitative experiments demonstrate the effectiveness and efficiency of our MooD, while ablation studies reveal the key design factors behind performance.

\end{itemize}

%% file: Sec/2_related.tex
\section{Related Work}
\label{sec:related}

\subsection{Computational Emotion Modeling}
In psychology, human emotions are typically modeled from two perspectives, \textit{i.e.}, CES and DES. CES (\textit{e.g.}, Mikels~\cite{mikels2005emotional}, Plutchik~\cite{plutchik}) represents emotions as discrete categories, whereas DES models them in continuous spaces such as Valence-Arousal (–Dominance)~\cite{circumplex_model}. Despite the greater expressiveness of DES, most image affective computing methods still adopt CES. For example, Wu et al.~\cite{emotions} analyzed the emotions of short videos based on Plutchik’s model. Such discrete formulations introduce discontinuities and limit fine-grained affect modeling. 
To overcome this limitation, DES-based methods represent emotions continuously, enabling finer-grained control, as exemplified by EmotiCrafter~\cite{emoticrafter}.

However, most methods in AIE remain CES-driven, and modeling emotions directly in VA space is still largely unexplored. The reliance on discrete categories limits continuity and controllability of emotional expressions. Motivated by this, we formulate AIE in the VA space for smoother and more fine-grained manipulation.

\subsection{Diffusion-based Image Editing}
Diffusion models have become a dominant paradigm for image editing due to their strong generative capability. Built upon text-to-image models~\cite{sd15,sdxl}, Prompt-to-Prompt~\cite{hertz2022prompttopromptimageeditingcross} and InstructPix2Pix~\cite{ip2p} enable language-guided editing through cross-attention control and instruction tuning. Recent methods further introduce structural or visual conditions for improved controllability. ControlNet~\cite{controlnet} uses spatial conditions to preserve structure, while IP-Adapter~\cite{ip-adapter} injects reference-image features for visual transfer. However, these conditions generally complement textual guidance, which remains important for constraining editing intent.

This observation is crucial for our VA-driven AIE. Relying solely on reference images may overemphasize visual transfer while lacking explicit emotional constraints. We thus claim visual conditions alone are insufficient for reliable affective editing. Motivated by this, we propose ASP-Net to project affective visual features from a reference image into adaptive CLIP text embeddings, injecting emotion-relevant cues into the diffusion model in a semantically aligned manner.

\begin{figure*}
    \centering
    \includegraphics[width=1.0\textwidth]{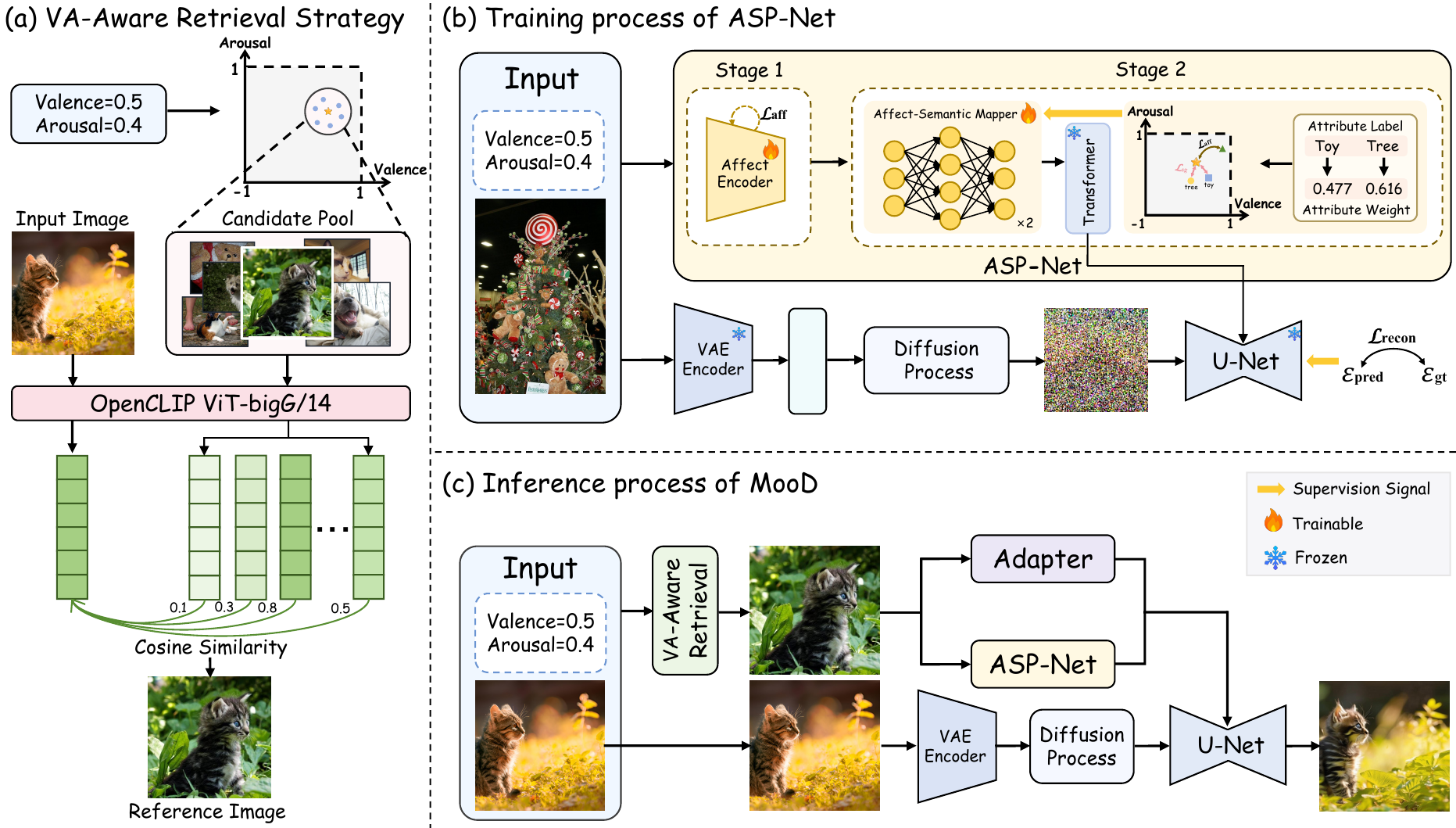}
    \caption{The overall illustration of our MooD framework. \textbf{(a)} The VA-Aware retrieval strategy selects reference images via VA-space filtering and OpenCLIP ViT-bigG/14 similarity. \textbf{(b)} The ASP-Net learns affective representations and maps them into the CLIP space through a two-stage training scheme, enabling emotion-aware conditioning. \textbf{(c)} The inference pipeline for emotion-aligned image editing given a source image and target VA values.}
    \label{fig:framework}
    \vspace{-1.5em}
\end{figure*}

\subsection{Affective Image Editing}

Early AIE methods focus on global appearance manipulation (\textit{e.g.}, color and style transfer) to evoke emotions. For instance, Peng et al.~\cite{color_transfer_3} transfer color templates and texture statistics from reference images, while Fu et al.~\cite{clva} manipulate the artistic style of content images according to textual instructions. However, these are limited to low-level modifications. Recent works shift toward semantic-level manipulation. EmoEdit~\cite{emoedit} introduces emotion adapters to modify or insert scene elements, and Make Me Happier~\cite{make_me_happier} adopts a dual-branch diffusion framework to preserve structure while injecting emotions. AIM-Bench~\cite{aimbench} explores hierarchical affect control, while AffectDelta~\cite{affectdelta} models emotion-distribution transitions. Meanwhile, advances in LLMs and MLLMs enable new paradigms. EmoAgent~\cite{emoagent} employs a multi-agent framework for complex editing. Despite improved semantic controllability, these methods often rely on increasingly complex inference pipelines.

Despite the above advances, existing methods still suffer from two limitations. First, many approaches rely on computationally expensive components, which limit practical efficiency and deployability. Second, most methods formulate affective control through discrete categories or language instructions, making it difficult to achieve fine-grained emotional manipulation. In contrast, we propose a lightweight framework that directly takes VA values as control signals, enabling efficient and fine-grained AIE without heavy pipelines.
 

%% file: Sec/3_method.tex
\section{Methodology}
\label{sec:method}

\subsection{Overview}

Given a source image $I_s$ and a target emotion $(v_t, a_t) \in [-1, 1]^2$, AIE aims to inject or modify emotion-relevant visual elements in $I_s$ to match $(v_t, a_t)$, while preserving the remaining content of $I_s$. Formally, we define this problem as learning a mapping, \textit{i.e.},
\begin{equation}
\mathcal{F}:(I_s, v_t, a_t) \rightarrow I_t,
\end{equation}
where $I_t$ satisfies both content fidelity to $I_s$ and affective consistency with $(v_t, a_t)$.

In this paper, we introduce MooD, an efficient dual-branch AIE framework, designed to guide the editing process from two complementary perspectives: transferring visual affective cues and enforcing semantic constraints. As illustrated in Fig.~\ref{fig:framework}~(c), given $I_s$ and $(v_t, a_t)$, we first retrieve a reference image $I_r$ using our VA-Aware retrieval strategy (Fig.~\ref{fig:framework}~(a)), which can be defined as:
\begin{equation}
I_r = \mathcal{R}(I_s, v_t, a_t),
\end{equation}
where $I_r$ is affectively aligned with $(v_t, a_t)$ while remaining semantically consistent with $I_s$. The resulting $I_r$ is then processed by two complementary branches, capturing both visual affective cues and high-level semantic guidance, which can be formulated as:
\begin{align}
c_v &=\mathcal{A}(I_r), \\
c_s &=\mathcal{P}(I_r),
\end{align}
where $\mathcal{A}(\cdot)$ refers to an image adapter that injects emotion-relevant visual patterns into the U-Net backbone via cross-attention. In parallel, $\mathcal{P}(\cdot)$ (\textit{i.e.}, our ASP-Net) projects $I_r$ into a CLIP text embedding space that is adaptively aligned with its emotional expression, providing perception-enhanced guidance for fine-grained emotional control while remaining compatible with SDXL’s text-conditioning mechanism.

Under the collaborative guidance of $(c_v, c_s)$, the diffusion process starts from the initial state $I_s$ for generation, which can be formulated as:
\begin{equation}
I_t = \mathcal{G}(I_s, c_v, c_s),
\end{equation}
where $\mathcal{G}(\cdot)$ denotes the SDXL generator. Notably, our ASP-Net is trained under the self-reconstruction paradigm to learn the mapping between visual features and the CLIP semantic space, as illustrated in Fig.~\ref{fig:framework}~(b). The detailed design of ASP-Net is presented in Sec.~\ref{subsec:3.3}.

\subsection{VA-Aware Retrieval Strategy}

Existing diffusion-based text-to-image models typically lack the ability to interpret numerical VA values. To bridge the gap between VA representations and visual semantics, we introduce a VA-Aware retrieval strategy. Specifically, we retrieve a reference image $I_r$ that provides explicit visual guidance for the target $(v_t,a_t)$, thereby grounding the numerical affective condition in concrete visual semantics. Retrieval is performed over a pre-constructed VA-annotated image database $\mathcal{D}=\{(I_i,v_i,a_i)\}_{i=1}^N$.
An ideal $I_r$ should satisfy two criteria: (i) strong affective alignment with $(v_t,a_t)$ to support accurate emotion transfer, and (ii) semantic compatibility with $I_s$ to avoid visual conflicts or semantic discontinuities. Since relying on either criterion alone may fail to satisfy both requirements, we adopt a two-stage retrieval strategy that jointly considers affective and semantic consistency.

First, we filter candidates in $\mathcal{D}$ by constructing a neighborhood centered at $(v_t, a_t)$ with radius $\tau$:
\begin{equation}
\mathcal{C}(v_t, a_t)=\left\{I_i\in\mathcal{D}:\left\|(v_i, a_i)-(v_t, a_t)\right\|_2<\tau\right\}.
\end{equation}

If $\mathcal{C}(v_t,a_t)$ is empty, it is instead formed by the $K=200$ samples closest to $(v_t,a_t)$ in the VA space.

Then, within the candidate pool $\mathcal{C}(v_t, a_t)$, cosine similarity is computed between each candidate image and $I_s$ using visual features extracted by an OpenCLIP ViT-bigG/14 encoder~\cite{clip}. The most similar sample is selected as the final $I_r$:
\begin{equation}
I_r=\underset{I_i\in\mathcal{C}(v_t, a_t)}{\arg\max}~\frac{\phi(I_i)^\top\phi(I_s)}{\left\|\phi(I_i)\right\|_2\cdot\left\|\phi(I_s)\right\|_2},
\end{equation}
where $\phi(\cdot)$ refers to the image encoder.

\subsection{Affective Semantic Projection Network (ASP-Net)}
\label{subsec:3.3}
We introduce image adapters to transfer visual features from reference images for affective editing, which can lead to artifacts. To enable more selective utilization of emotion-relevant elements, we propose ASP-Net, which comprises an affect-aware visual representation module and an affect-semantic mapper.

\begin{table}[t!]
\centering
\setlength{\abovecaptionskip}{0.4em}
\renewcommand{\arraystretch}{0.98}
\setlength{\arrayrulewidth}{0.22pt}
\caption{Annotation validation against human annotations.}
\resizebox{0.8\linewidth}{!}{
\begin{tabular}{cccc}
\toprule
Dimension & Pearson r & CCC    & MAE \\ \midrule[0.22pt]
Valence   & 0.914     & 0.913  & 0.138   \\
Arousal   & 0.837     & 0.836  & 0.165\\
\bottomrule
\end{tabular}
}
\label{MLLM_compare}
\vspace{-1.5em}
\end{table}

\begin{table*}[t!]
\centering
\setlength{\tabcolsep}{3pt}
\small
\caption{The statistical distribution of our AffectSet dataset in the VA space and discrete emotion categories.}

\resizebox{\textwidth}{!}{
\begin{tabular}{cccccccccccccc}
\toprule
\multirow{2}{*}{\makecell[c]{Metric}}
& \multirow{2}{*}{Total}
& \multicolumn{4}{c}{VA Quadrants}
& \multicolumn{8}{c}{Discrete Emotion Categories} \\
\cmidrule(lr){3-6} \cmidrule(l){7-14}
& & V+ A+ & V+ A- & V- A+ & V- A-
& Amusement & Awe & Contentment & Excitement
& Anger & Disgust & Fear & Sadness \\
\midrule
Count
& 63387
& 38604 & 4011 & 13088 & 7684
& 10963 & 7207 & 10287 & 13561
& 5400 & 4332 & 4761 & 6876 \\
\bottomrule
\end{tabular}
}
\label{dataset}
\end{table*}

\subsubsection{Affect-Aware Visual Representation}

Generic visual encoders (\textit{e.g.}, CLIP) are optimized for semantic discriminability, and thus their feature spaces do not explicitly encode emotional similarity. For instance, a dark forest and a dimly lit room may evoke similar emotions despite differing semantics, while the same forest under varying lighting conditions remains semantically consistent but exhibits distinct emotional states. As a result, we construct an affect encoder based on ResNet-50~\cite{resnet}. By incorporating affective supervision, our encoder reshapes the visual feature space to better capture emotional similarity, thereby producing affect-aligned representations that facilitate subsequent cross-modal mapping. 

Following the VA modeling paradigm, we introduce an encoder-level affect consistency loss:
\begin{equation}
\mathcal{L}_{\text{aff}}^E = \|(\hat{v_E}, \hat{a_E}) - (v, a)\|_2^2,
\end{equation}
where $(\hat{v}_E, \hat{a}_E)$ and $(v, a)$ denote the predicted and ground-truth VA values, respectively. After training, the encoder is frozen to decouple feature learning from subsequent semantic projection, preventing drift during joint optimization. We extract features from the penultimate stage as the affect feature $f_{\text{emo}} \in \mathbb{R}^{d_L}$, whose dimensionality is aligned with the CLIP ViT-L/14 text embedding space used in SDXL.

\begin{table*}[t!]
\centering
\caption{The quantitative comparison of MooD with state-of-the-art image editing methods. Gray rows indicate ULMs and GAN-based approaches. TMP denotes Time per Megapixel (Time/MP). Params (T / Tot.) represent the number of trainable parameters and total parameters, respectively. The best and second-best results among comparable methods are highlighted in bold and \underline{underlined}, respectively.}
\label{tab:quantitative_comparison}
\vspace{-0.3em}
\setlength{\tabcolsep}{3.5pt}
\small
\renewcommand{\arraystretch}{1.21}
\resizebox{1.0\textwidth}{!}{
\begin{tabular}{@{} lccccccccc @{}}
\toprule
Method & Org. & Params (T / Tot.) & TMP $\downarrow$ & PSNR $\uparrow$ & SSIM $\uparrow$ & LPIPS $\downarrow$ & CLIP-I $\uparrow$ & V-Err $\downarrow$ & A-Err $\downarrow$ \\
\midrule
EmoEdit~\cite{emoedit} & SZU & 99.30 M / 1.59 B & \underline{16.196} & 15.136 & \textbf{0.556} & 0.321 & \underline{0.813} & 0.578 & 0.308 \\
Make Me Happier~\cite{make_me_happier} & NTU & 0.91 B ~/ 1.00 B & 560.907 & 16.406 & \underline{0.541} & \underline{0.311} & 0.763 & 0.695 & 0.344 \\
SDEdit~\cite{sdedit} & Stanford & ~~~~-~~~~~/ 1.06 B & 350.031 & \underline{16.830} & 0.442 & 0.363 & 0.683 & 0.596 & 0.329 \\
ControlNet~\cite{controlnet} & Stanford & 0.36 B~ / 1.42 B & 24.478 & 9.444 & 0.211 & 0.554 & 0.750 & 0.839 & \underline{0.306} \\
P2P‐Zero~\cite{p2p_zero} & CMU & ~~~~-~~~~~/ 1.06 B & 39.018 & 13.760 & 0.420 & 0.546 & 0.685 & \underline{0.567} & 0.581 \\
InstructPix2Pix~\cite{ip2p} & Berkeley & 0.86 B~ / 1.06 B & 31.670 & 11.020 & 0.399 & 0.481 & 0.318 & 0.712 & 0.485 \\
\cmidrule{1-10}
\color{gray} Step1X~\cite{step1x} & \color{gray} StepFun & \color{gray} 12.40 B~/ 19.00 B & \color{gray} 24.092 & \color{gray} 15.281 & \color{gray} 0.598 & \color{gray} 0.210 & \color{gray} 0.892 & \color{gray} 0.662 & \color{gray} 0.415 \\
\color{gray} DreamOmni2~\cite{dreamomni2} & \color{gray} CUHK & \color{gray} 0.72 B~~/ 19.00 B & \color{gray} 24.597 & \color{gray} 17.720 & \color{gray} 0.601 & \color{gray} 0.280 & \color{gray} 0.902 & \color{gray} 0.720 & \color{gray} 0.385 \\
\cmidrule{1-10}
\color{gray} AIF~\cite{style_transfer_1} & \color{gray} PKU & \color{gray} 29.92 M / 43.25 M & \color{gray} 0.027 & \color{gray} 14.668 & \color{gray} 0.576 & \color{gray} 0.379 & \color{gray} 0.825 & \color{gray} 0.792 & \color{gray} 0.400 \\
\color{gray} CLVA~\cite{clva} & \color{gray} UCSB & \color{gray} 154.81 M / 171.26 M & \color{gray} 0.526 & \color{gray} 13.097 & \color{gray} 0.518 & \color{gray} 0.305 & \color{gray} 0.842 & \color{gray} 0.737 & \color{gray} 0.564 \\
\hline

\textbf{MooD (Ours)} & ZZU & 38.78 M / 4.45 B & \textbf{4.500} & \textbf{17.509} & 0.518 & \textbf{0.287} & \textbf{0.825} & \textbf{0.537} & \textbf{0.287} \\
\bottomrule
\end{tabular}
}
\vspace{-1.5em}
\end{table*}

\subsubsection{Affect-Semantic Mapper}

The affect feature $f_{\text{emo}}$ lies in a continuous affect space misaligned with the text embedding space in SDXL~\cite{sdxl}, making it incompatible with the text encoder. Therefore, we introduce an affect-semantic mapper that projects affect features into the CLIP text space. EmoGen~\cite{emogen} shows that a single affect category often corresponds to multiple semantic clusters in the CLIP space, indicating a one-to-many relationship between affective states and semantic expressions. This suggests that representing affect with a single embedding is insufficient to capture semantic diversity. We therefore adopt a multi-token representation to capture diverse affective semantics.

The mapper is implemented as an MLP with GELU activation, producing a set of tokens $s \in \mathbb{R}^{B \times N \times d}$. Given that SDXL employs dual text encoders (CLIP ViT-L/14 and OpenCLIP ViT-bigG/14), we further employ two mapping branches to generate local and global semantic embeddings, \textit{i.e.},
\begin{align}
s_L = MLP_L(f_{\text{emo}}) \in \mathbb{R}^{N \times d_L}, \\
s_G = MLP_G(f_{\text{emo}}) \in \mathbb{R}^{N \times d_G},
\end{align}
where $d_G$ is aligned with the OpenCLIP ViT-bigG/14 text embedding space utilized in SDXL. Subsequently, $s_L$ and $s_G$ respectively replace the embeddings of $N$ placeholder tokens at the input embedding layers of the two text encoders, thereby incorporating multi-token affect representations into SDXL.

\subsubsection{Training Objectives with Semantic Anchoring}

ASP-Net is trained in a self-reconstruction paradigm (Fig.~\ref{fig:framework}~(b)). The overall objective is:
\begin{equation}
\mathcal{L} = \mathcal{L}_{\text{rec}} + \alpha \mathcal{L}_{\text{aff}}^M + \beta \mathcal{L}_{\text{sg}},
\end{equation}
where $\mathcal{L}_{\text{rec}}$ is the diffusion denoising loss and $\mathcal{L}_{\text{aff}}^M$ is the mapper-level affect consistency loss preserving affective information during semantic projection.
To enhance semantic alignment, we introduce a semantic grounding loss $\mathcal{L}_{\text{sg}}$.

\noindent \textbf{Semantic grounding loss $\mathcal{L}_{\text{sg}}$.} 
We leverage the rich object attribute annotations provided by EmoSet as semantic supervision signals. A straightforward approach is to compute similarity between predicted embeddings and attribute CLIP embeddings. However, a single image may convey diverse attributes with unequal relevance to the target emotion. Assigning equal weights to all attributes may cause the mapper to allocate capacity to emotion-irrelevant semantics, thereby weakening affect modeling.
To mitigate this issue, we design a Mahalanobis-distance-based dynamic attribute weighting mechanism. Specifically, image attributes are represented as a multi-hot vector $\mathbf{y}$, with annotated attributes as positive labels and the remaining retained attributes as negative labels. Prediction probabilities are computed via cosine similarity between semantic and attribute text embeddings with BCE supervision. We then assign weights according to the Mahalanobis distance between the target VA value and each attribute’s distribution in the continuous VA space:

\vspace{-1em}
\begin{equation}
w_k(v,a) = \exp\left(-\frac{\gamma}{2} \, 
((v,a) - \boldsymbol{\mu}_k)^\top \boldsymbol{\Sigma}_k^{-1} 
((v,a) - \boldsymbol{\mu}_k)\right),
\end{equation}
where $\boldsymbol{\mu}_k$ and $\boldsymbol{\Sigma}_k$ denote the mean and covariance of attribute $k$ in the VA space. $\gamma$ is a scaling factor controlling the sharpness of the weighting (set to 1 by default), under which $w_k(v,a)$ takes the form of a Gaussian kernel determined by the natural statistics of the VA distribution. 

To ensure numerical stability, each empirical covariance matrix is regularized as $\widetilde{\boldsymbol{\Sigma}}_k=\boldsymbol{\Sigma}_k+\lambda\mathbf I_2$, where $\lambda=10^{-4}$, before computing the Mahalanobis distance. The final attribute loss can be formulated as:
\begin{equation}
\mathcal{L}_{\text{sg}} = \sum_{k=1}^{K} w_k(v,a) \cdot \text{BCE}_k,
\end{equation}
where $K$ is the total number of attributes and $\text{BCE}_k$ denotes the binary cross-entropy loss for attribute $k$.

\subsection{AffectSet Dataset}

To facilitate model training and evaluation, we construct the AffectSet dataset with continuous VA annotations. Concretely, we build upon EmoSet~\cite{emoset} and extend its discrete emotion labels into the VA space by employing Qwen2.5-VL-72B~\cite{Qwen2.5-VL}, resulting in a total of 63,387 samples. To assess annotation quality, we divide both Valence and Arousal into five intervals, forming 25 VA strata, and sample up to 20 images from each stratum, yielding 465 images for human evaluation following~\cite{sampling}. For a stratum containing fewer than 20 images, we include all available samples. As shown in Table~\ref{MLLM_compare}, the results demonstrate strong agreement between model-generated annotations and human judgments, indicating the validity of the constructed VA labels. Detailed dataset statistics are summarized in Table~\ref{dataset}. Furthermore, we model the distribution of each attribute category in the VA space as a Gaussian $\mathcal{N}(\boldsymbol{\mu}_a, \boldsymbol{\Sigma}_a)$ for categories with sufficient samples (more than 30 instances). These distributions are later incorporated into the semantic grounding loss $\mathcal{L}_{\text{sg}}$ to enable dynamic, attribute-aware weighting.

%% file: Sec/4_expers.tex
\section{Experiments}
\label{sec:experiment}

\subsection{Experimental Settings}

\noindent \textbf{Implementation Details.} MooD is implemented in PyTorch 2.3.0~\cite{pytorch} with AdamW~\cite{adamw}. The Affect Encoder is trained for 100 epochs on a single NVIDIA RTX 4090D GPU (batch size=128, lr=$1\times10^{-3}$, weight decay=$1\times10^{-4}$), using random resized crops of $224\times224$ for training and resize-center crops for validation. The resulting 768-dimensional penultimate-layer features are fed into the Affect-Semantic Mapper, which is trained for 20 epochs on 6 $\times$ RTX 4090D GPUs (batch size=1 per GPU, lr=$3\times10^{-5}$, weight decay=$1\times10^{-2}$, 300-step warmup) using the same validation preprocessing. Images are resized to $1024\times1024$ and normalized to $[-1,1]$ before SDXL VAE encoding. AffectSet is split 8:2 for training and validation. Evaluation uses the EmoEdit inference set, with discrete emotions mapped to continuous VA values using the NRC-VAD Lexicon~\cite{mohammad2025nrc}. All comparison methods follow their official settings and are evaluated on a single NVIDIA RTX 4090 GPU, except Step1X and DreamOmni2, which use an RTX PRO 6000 GPU due to their larger model sizes.

\noindent \textbf{Evaluation Metrics.} To evaluate AIE, we analyze three aspects: inference efficiency, image quality, and affective controllability. For efficiency, we report \textbf{Time/MP}, measuring total inference time (seconds) per million pixels. For image quality, following~\cite{emoedit}, we use \textbf{PSNR} for pixel fidelity, \textbf{SSIM} for structural similarity, \textbf{LPIPS} for perceptual distance in deep feature space, and \textbf{CLIP-I} to measure semantic consistency between the source and edited images. For affective controllability, we compute \textbf{Valence Error (V-Err)} and \textbf{Arousal Error (A-Err)}, defined as the $\ell_1$ distance between predicted VA values of edited images and target emotions.

\begin{figure*}[t!]
    \centering
    \includegraphics[width=\linewidth]{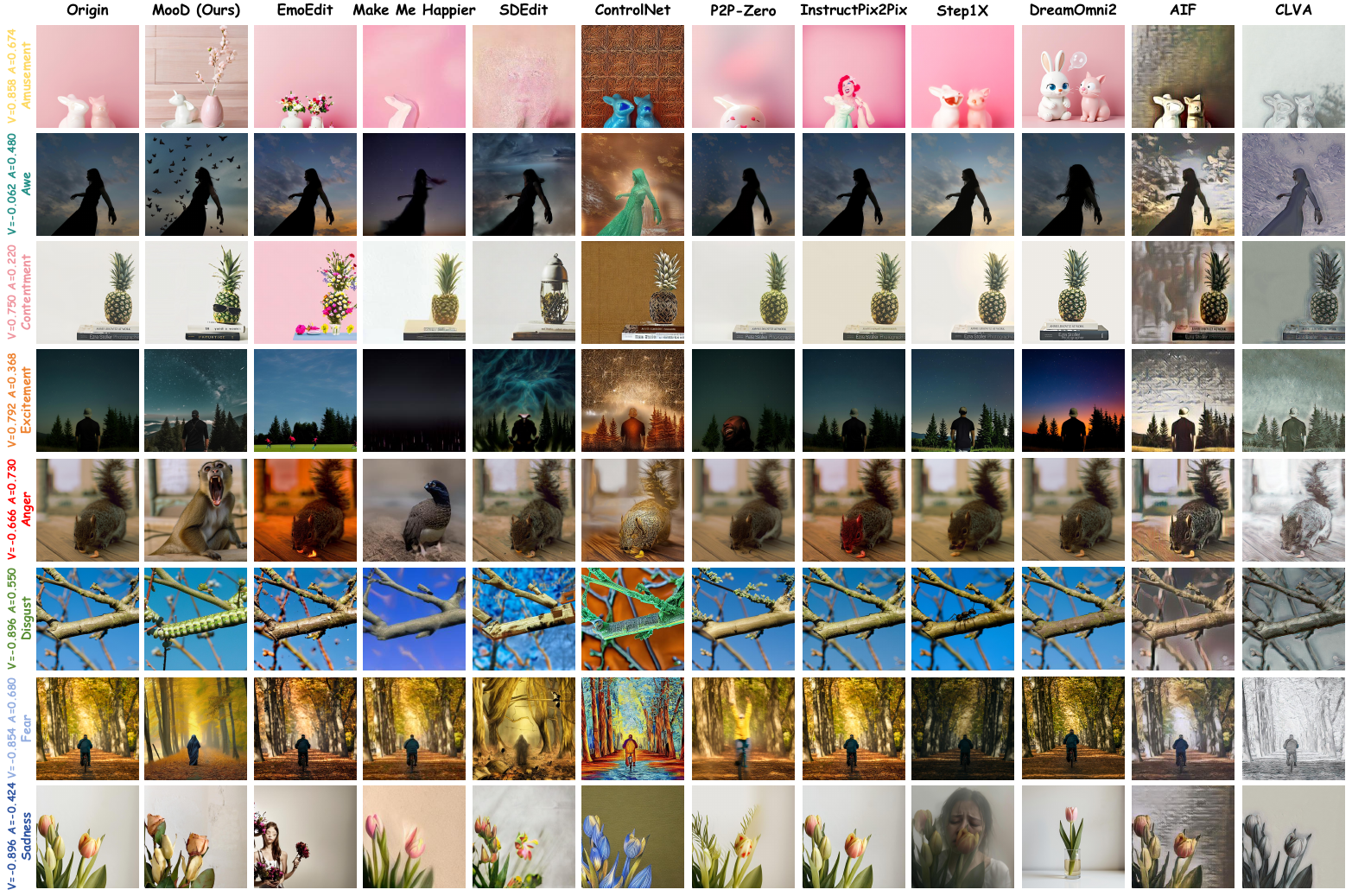}
    \caption{Qualitative Comparison with state-of-the-art methods. Our MooD outperforms existing approaches in both affective controllability and structural content preservation.}
    \vspace{-0.8em}
    \label{fig:comparison}
\end{figure*}

\begin{figure*}[t!]
    \centering
    \includegraphics[width=\linewidth]{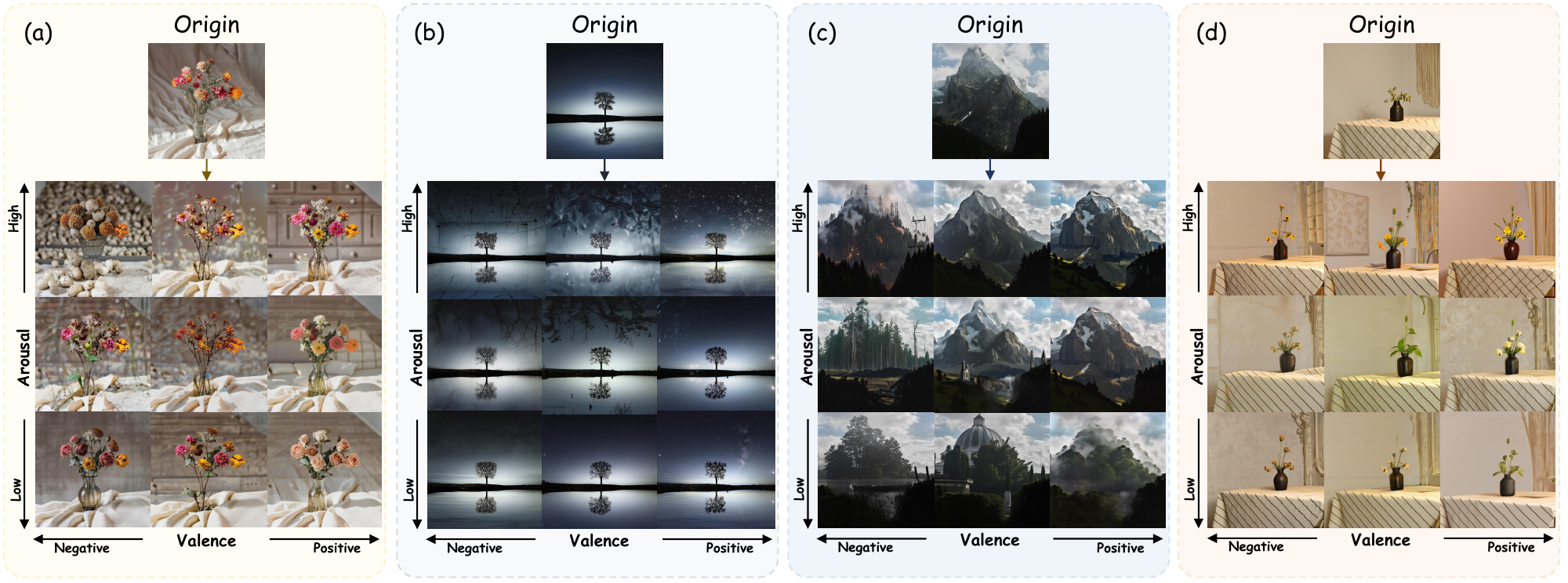}
    \caption{The visualization of smooth affective transitions achieved by our MooD in the VA space. Each row (left to right) reflects a gradual shift in valence from negative to positive, while each column (bottom to top) exhibits a steady increase in arousal.}
    \label{fig:va_sweep_test}
    \vspace{-1.4em}
\end{figure*}

\subsection{Main Results}
To evaluate MooD, we compare it with representative state-of-the-art methods across four categories:
(i) general image editing methods, including SDEdit~\cite{sdedit}, ControlNet~\cite{controlnet}, P2P-Zero~\cite{p2p_zero}, and InstructPix2Pix~\cite{ip2p};
(ii) AIE methods, including EmoEdit~\cite{emoedit} and Make Me Happier~\cite{make_me_happier};
(iii) unified large models (ULMs), including Step1X~\cite{step1x} and DreamOmni2~\cite{dreamomni2}; and
(iv) style-transfer-based AIE methods, including AIF~\cite{style_transfer_1} and CLVA~\cite{clva}.

\begin{figure*}[t!]
    \centering
    \includegraphics[width=\linewidth]{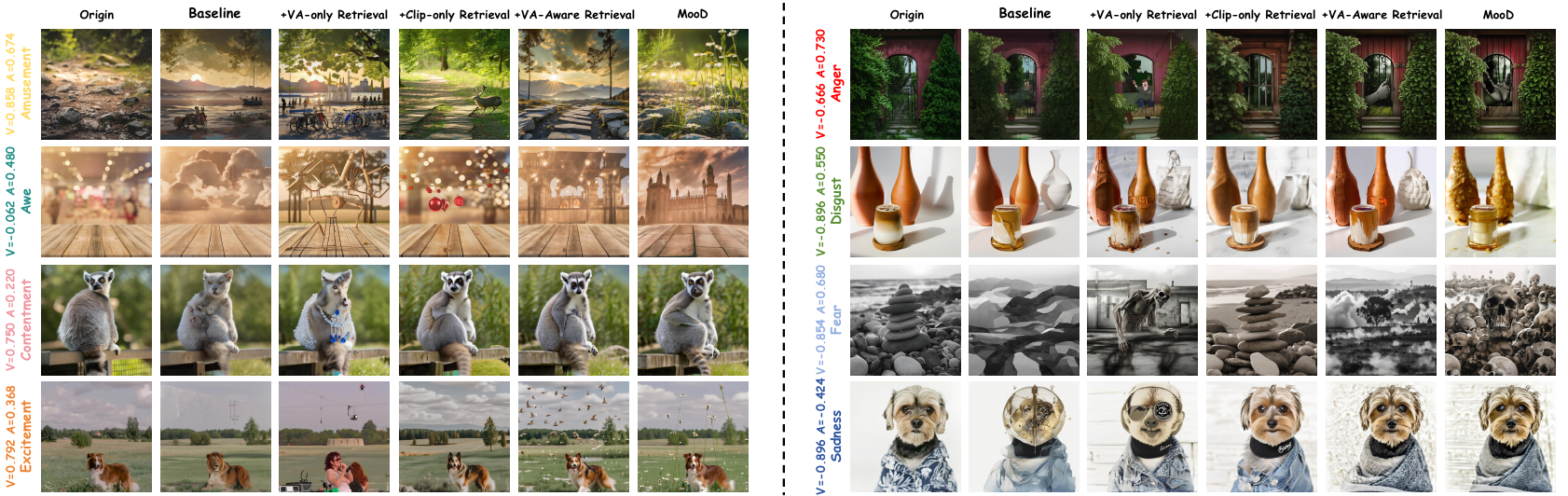}
    \caption{The qualitative ablation results of key components in our MooD, demonstrating that both the VA-Aware retrieval strategy and ASP-Net are indispensable.}
    \vspace{-1.5em}
    \label{fig:ablation}
\end{figure*}

\subsubsection{Quantitative Comparisons}
As shown in Table~\ref{tab:quantitative_comparison}, MooD achieves the lowest Valence Error (0.5366) and Arousal Error (0.2867), demonstrating effective fine-grained affective control in the continuous VA space. It also achieves the best PSNR, LPIPS, and CLIP-I among methods of comparable scale, outperforming the second-best method by 4.034\%, 7.717\%, and 1.476\%, respectively. The slightly lower SSIM may result from luminance and contrast changes introduced by reference-based content transfer. Compared with ULMs, MooD achieves comparable performance with substantially fewer parameters, while style-transfer methods preserve structure but lack explicit affect modeling. Moreover, MooD reaches 4.5 s/MP, substantially faster than diffusion-based methods (\textit{e.g.}, (i), (ii)), confirming its overall efficiency.

\begin{table}[t]
\centering
\caption{Ablation study of MooD. Starting from the baseline, we progressively introduce VA-only retrieval strategy, CLIP-only retrieval strategy, and the proposed VA-Aware retrieval strategy, followed by the full model with ASP-Net.}
\label{ablation_results}
\resizebox{\columnwidth}{!}{
\begin{tabular}{lccccc}
\toprule
Method & PSNR $\uparrow$ & SSIM $\uparrow$ & CLIP-I $\uparrow$ & V-Err $\downarrow$ & A-Err $\downarrow$ \\
\midrule
Baseline & \textbf{18.497} & \textbf{0.544} & 0.651 & 0.753 & 0.386 \\
+ VA-only Retrieval & 16.887 & 0.494 & 0.491 & \underline{0.581} & \underline{0.328} \\
+ CLIP-only Retrieval & 17.466 & \underline{0.518} & \underline{0.785} & 0.764 & 0.379 \\
+ VA-Aware Retrieval & 17.371 & 0.511 & 0.709 & 0.670 & 0.354 \\
+ VA-Aware Retrieval + ASP-Net & \underline{17.509} & \underline{0.518} & \textbf{0.825} & \textbf{0.537} & \textbf{0.287} \\
\bottomrule
\end{tabular}
}
\vspace{-1.2em}
\end{table}

\begin{figure}[t!]
    \centering
    \includegraphics[width=\linewidth]{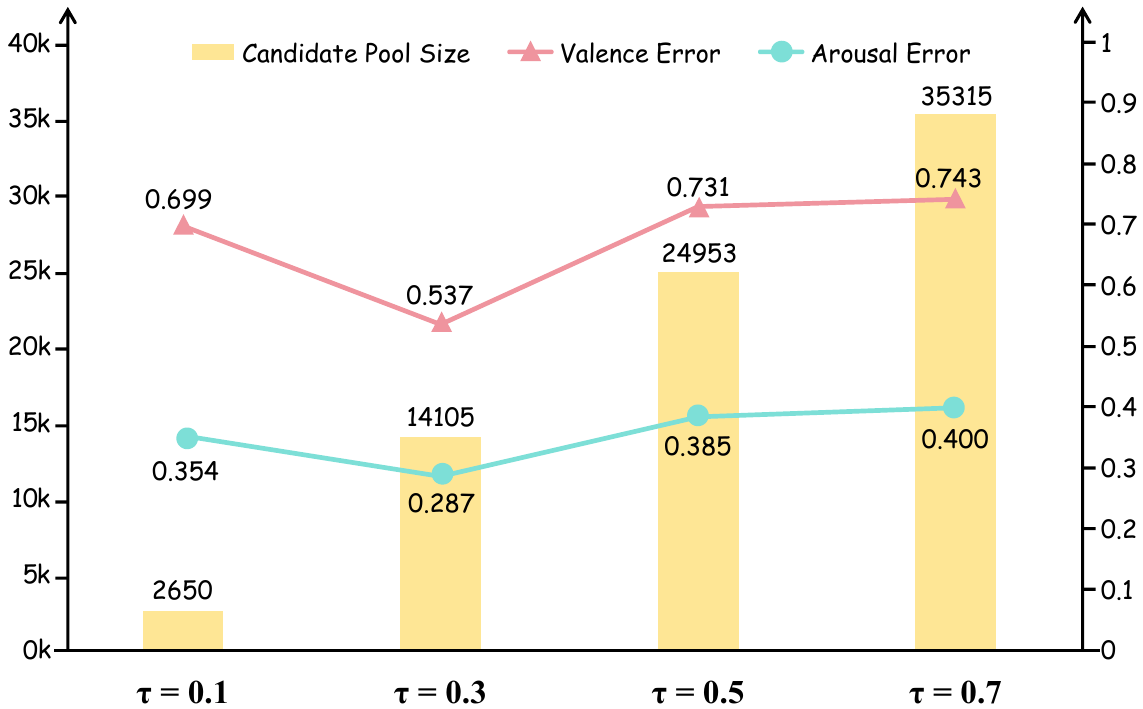}
    \caption{Effect of $\tau$ on our VA-Aware retrieval strategy.}
    \label{fig:tau_analysis}
    \vspace{-1.8em}
\end{figure}

\subsubsection{Qualitative Comparisons}

\textbf{Qualitative Comparison with SOTA Methods.} We present qualitative results in Fig.~\ref{fig:comparison}. MooD balances content preservation and emotion expression, selectively modifying emotion-relevant elements while keeping other regions largely unchanged, and shows stronger emotional elicitation than prior AIE methods. For example, for the anger target, EmoEdit~\cite{emoedit} generates fire-related content, while Make Me Happier~\cite{make_me_happier} produces semantically ambiguous objects. General editing methods offer limited emotional control, often introducing context-inconsistent changes or weak affect. Step1X~\cite{step1x} and DreamOmni2~\cite{dreamomni2} preserve semantics well but yield conservative edits with occasional unnecessary structural changes. AIF~\cite{style_transfer_1} and CLVA~\cite{clva} rely on style-level transformations, lacking fine-grained affect modeling.

\noindent \textbf{Affective Transition Capability of MooD.}~To further evaluate MooD, we present four representative samples across the VA space in Fig.~\ref{fig:va_sweep_test}. Across diverse scenes, MooD achieves stable and continuous affective control. Increasing Valence shifts the appearance from dim and suppressed to bright and warm, as shown by shifts from moldy fungi to blooming flowers and from burning mountains to clear landscapes. Increasing Arousal enriches structures, textures, and perceptual dynamics, as observed in the starry sky and floral examples. Under joint two-dimensional control, Valence and Arousal remain approximately disentangled in most regions and vary smoothly and monotonically along their corresponding directions. These results demonstrate the fine-grained control of MooD in the continuous VA space and its effectiveness for AIE.

\subsection{Ablation Study}

\begin{table}[t]
\centering
\caption{Impact of ASP-Net token number $N$.}
\label{n_results}
\resizebox{\columnwidth}{!}{
\begin{tabular}{lccccc}
\toprule
Setting & PSNR $\uparrow$ & SSIM $\uparrow$ & CLIP-I $\uparrow$ & V-Err $\downarrow$ & A-Err $\downarrow$ \\
\midrule
$N=2$ & 17.379 & 0.509 & 0.684 & 0.736 & 0.376 \\
$N=3$ & \underline{17.491} & \underline{0.517} & \underline{0.817} & \underline{0.620} & \underline{0.308} \\
$N=4$ & \textbf{17.509} & \textbf{0.518} & \textbf{0.825} & \textbf{0.537} & \textbf{0.287} \\
$N=5$ & 17.397 & 0.514 & 0.718 & 0.703 & 0.318 \\
\bottomrule
\end{tabular}
}
\vspace{-0.6em}
\end{table}

\begin{table}[t]
\centering
\caption{Effect of loss weights $\alpha$ and $\beta$ on performance.}
\label{tab:loss_weight_results}
\resizebox{\columnwidth}{!}{
\begin{tabular}{ccc ccccc}
\toprule
\multicolumn{3}{c}{Setting}
& \multirow{2}{*}{PSNR $\uparrow$}
& \multirow{2}{*}{SSIM $\uparrow$}
& \multirow{2}{*}{CLIP-I $\uparrow$}
& \multirow{2}{*}{V-Err $\downarrow$}
& \multirow{2}{*}{A-Err $\downarrow$} \\
\cmidrule(lr){1-3}
$\alpha$ & $\beta$ & Variant & & & & & \\
\midrule
- & 0.5 & w/o $\mathcal{L}_{\text{aff}}^M$ & \underline{17.410} & 0.508 & 0.676 & 0.726 & 0.370 \\
0.1 & - & w/o $\mathcal{L}_{\text{sg}}$ & 17.356 & 0.509 & \underline{0.703} & 0.686 & 0.347 \\
1.0 & 0.5 & -- & 17.295 & 0.507 & 0.694 & \underline{0.674} & \underline{0.329} \\
0.1 & 1.0 & -- & 17.397 & \underline{0.510} & 0.698 & 0.741 & 0.360 \\
0.1 & 0.5 & Ours & \textbf{17.509} & \textbf{0.518} & \textbf{0.825} & \textbf{0.537} & \textbf{0.287} \\
\bottomrule
\end{tabular}
}
\vspace{-1.5em}
\end{table}

\subsubsection{Quantitative Analysis}

To assess the contributions of key components and hyperparameter configurations in MooD, a systematic ablation study is conducted.

\begin{table*}[t]
\centering
\color{black}
\caption{User study results. The first three metrics measure human-perceived VA alignment, while the last two evaluate structural preservation and affective faithfulness. Best results are in bold.}
\label{tab:user-study-results}
\small
\renewcommand{\arraystretch}{1.15}

\begingroup
\setlength{\tabcolsep}{5pt}
\begin{tabular*}{\textwidth}{@{\extracolsep{\fill}}lccccc@{}}
\toprule
Method
& Human V-Err $\downarrow$
& Human A-Err $\downarrow$
& Human VA-Err $\downarrow$
& Structure Preservation $\uparrow$
& Affective Faithfulness $\uparrow$ \\
\midrule

EmoEdit~\cite{emoedit}
  & $0.42 \pm 0.14$
  & $\mathbf{0.37 \pm 0.12}$
  & $0.40 \pm 0.10$
  & $2.75 \pm 1.42$
  & $3.17 \pm 0.72$ \\

Make Me Happier~\cite{make_me_happier}
  & $0.57 \pm 0.16$
  & $0.45 \pm 0.15$
  & $0.51 \pm 0.14$
  & $2.17 \pm 1.11$
  & $2.42 \pm 0.79$ \\

Step1X~\cite{step1x}
  & $0.50 \pm 0.15$
  & $0.38 \pm 0.12$
  & $0.44 \pm 0.11$
  & $3.75 \pm 1.48$
  & $3.58 \pm 1.00$ \\

MooD (Ours)
  & $\mathbf{0.35 \pm 0.12}$
  & $0.39 \pm 0.13$
  & $\mathbf{0.37 \pm 0.10}$
  & $\mathbf{4.42 \pm 0.79}$
  & $\mathbf{4.83 \pm 0.39}$ \\

\bottomrule
\end{tabular*}
\endgroup
\vspace{-1.5em}
\end{table*}

\noindent\textbf{Component contribution analysis.} As shown in Table~\ref{ablation_results}, directly using VA values as SDXL prompts yields the highest PSNR and SSIM but the largest VA Error, indicating limited understanding of VA-related emotional semantics. VA-only retrieval improves affective alignment at the cost of image quality, whereas CLIP-only retrieval better preserves visual content but fails to ensure the target affect. The proposed VA-Aware retrieval balances semantic fidelity and affective control. Further incorporating ASP-Net achieves the best overall performance, with the highest CLIP-I of 0.8252, competitive PSNR and SSIM, and reduced Valence and Arousal Errors of 0.5366 and 0.2867, respectively.

\noindent \textbf{Effect of the VA retrieval threshold $\tau$.} This threshold controls the candidate pool size, thereby balancing retrieval diversity and affective precision. As shown in Fig.~\ref{fig:tau_analysis}, when $\tau = 0.3$, the candidate pool reaches a moderate size, while Valence Error and Arousal Error attain the lowest values of 0.537 and 0.287, respectively. Further increasing $\tau$ expands the candidate pool but consistently reduces affective accuracy.

\noindent \textbf{Impact of ASP-Net token number.} Table~\ref{n_results} reports the effect of varying output token numbers in ASP-Net. A small token set ($N=2$) severely limits representational capacity, leading to the worst performance. Increasing $N$ enables richer and more fine-grained affective representations, yielding consistent improvements across all metrics and peaking at $N=4$. Further increasing the token number ($N=5$), however, results in performance degradation, likely caused by redundant or noisy semantic information.

\noindent \textbf{Effect of loss weights.} Table \ref{tab:loss_weight_results} presents an analysis of the influence of the weight $\alpha$ for {$\mathcal{L}_{\text{aff}}^M$ and the weight $\beta$ for $\mathcal{L}_{sg}$ on model performance. Removing $\mathcal{L}_{\text{aff}}^M$ severely degrades affect modeling, while removing $\mathcal{L}_{\text{sg}}$ (compared with $\alpha=0.1,\ \beta=0.5$) consistently harms all metrics, highlighting the necessity of semantic guidance. Moreover, increasing either weight degrades performance. This suggests that overly strong affective regression weakens visual and structural modeling, whereas excessive semantic constraints restrict affective flexibility.

\subsubsection{Qualitative Analysis}
Fig.~\ref{fig:ablation} qualitatively illustrates the contribution of each component. The baseline lacks affective understanding and produces ambiguous results. VA-only retrieval introduces affect cues but causes semantic drift (\textit{e.g.}, land to riverside and dog to human), whereas CLIP-only retrieval improves semantic consistency but poorly conveys target emotions, especially negative ones. VA-aware retrieval better balances affect and semantics but still suffers from structural hallucinations. Incorporating ASP-Net further improves structural fidelity and affective expression.

\begin{figure}[t!]
    \centering
    \includegraphics[width=\linewidth]{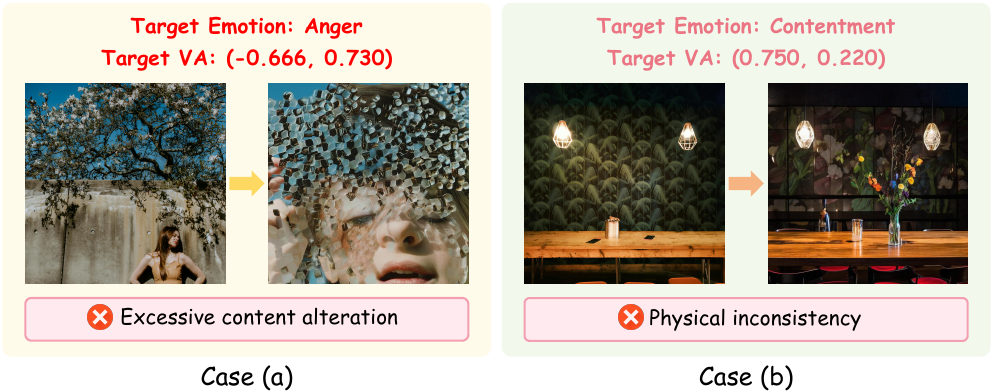}
    \caption{Representative failure cases of MooD.}
    \vspace{-1.5em}
    \label{fig:failure}
\end{figure}

\subsection{User Study}
We recruit 33 participants to compare MooD with EmoEdit, Make Me Happier, and Step1X through two complementary evaluation protocols, assessing both human-perceived VA alignment and comparative editing quality.

\noindent\textbf{Human-Perceived VA Alignment.}
Participants rate the valence and arousal of each anonymized result on nine-point scales, without target VA values or method identities. As shown in Table~\ref{tab:user-study-results}, MooD achieves the lowest Human V-Err of $0.35 \pm 0.12$ and Human VA-Err of $0.37 \pm 0.10$, indicating the best alignment with human affect perception. The slightly higher Human A-Err reflects the greater sensitivity of human arousal perception to complex visual and semantic cues.

\noindent\textbf{Editing Quality Comparison.}
For each source–target pair, participants are asked to compare four anonymized editing results produced by different methods for the same source image and target VA values. They rate each result on five‑point scales for content and structure preservation and affective faithfulness. As shown in Table~\ref{tab:user-study-results}, MooD obtains the highest scores of $4.42 \pm 0.79$ and $4.83 \pm 0.39$, outperforming Step1X by 0.67 and 1.25, respectively.

\subsection{Limitations}
Although MooD performs well, it still has several limitations. As shown in Fig.~\ref{fig:failure}, high-arousal editing may cause excessive structural and textural changes, while some results exhibit incorrect spatial relationships or physical inconsistencies. These issues arise because the model samples from learned visual distributions without explicit physical or spatial constraints. Improving physical plausibility remains an important direction for future work.

%% file: Sec/5_conclusions.tex
\section{Conclusion and Discussion}
\label{sec:discussion}

In this paper, we propose MooD, a new framework that enables efficient and fine-grained affective image editing (AIE)  by directly operating in the continuous Valence–Arousal (VA) space. By bridging affective values with visual semantics through VA-Aware retrieval and combining visual transfer with perception-enhanced semantic guidance, MooD provides an efficient solution for controllable AIE. Beyond performance improvements, our work highlights the importance of continuous affect modeling for capturing subtle emotional variations, offering a more expressive alternative to conventional discrete paradigms. We also contribute AffectSet, a VA-annotated dataset that facilitates future research in this direction.

Despite promising results, MooD relies on a pre-constructed affect-annotated image pool, which may limit diversity. The uneven VA distribution may also limit editing performance in sparse regions. Future work will explore more flexible affect representations, reduced reliance on external retrieval, and privacy-preserving dataset expansion~\cite{add_1}. MooD must be used with proper authorization and clear disclosure of AIE.